\documentclass[conference]{IEEEtran}
\IEEEoverridecommandlockouts
\usepackage{cite}
\usepackage{amsmath,amssymb,amsfonts}
\usepackage{algorithmic}
\usepackage{graphicx}
\usepackage{textcomp}
\usepackage{xcolor}
\usepackage{xspace}
\usepackage{subfig}
\usepackage{tabu}
\usepackage{tabularx}
\usepackage{multirow}
\usepackage{breakurl}
\usepackage[breaklinks]{hyperref}
\usepackage{algorithm}
\usepackage{algorithmic}
\usepackage{xspace}

\usepackage{newfloat}
\usepackage{listings}
\newcommand{\alg}{{FCT-GAN}\xspace}
\newcommand{\algg}{{FCT-GAN$_G$}\xspace}
\newcommand{\algd}{{FCT-GAN$_D$}\xspace}
\newcommand{\ctab}{{CTAB-GAN}\xspace}
\newcommand{\tablegan}{{Table-GAN}\xspace}

\newcommand{\ctgan}{{CT-GAN}\xspace}
\newcommand{\pategan}{{PATE-GAN}\xspace}
\newcommand{\itgan}{{IT-GAN}\xspace}
\newcommand{\tvae}{{TVAE}\xspace}
\newcommand{\ctabplus}{{CTAB-GAN+}\xspace}

\newcommand{\blocks}{{blocks}\xspace}

\author{

   \IEEEauthorblockN{Zilong Zhao\IEEEauthorrefmark{1}%\thanks{$^1$ This work has been partly supported by the IRS (Initialtive de Recherche Strat\'egique) program DATE.}
   ,  Robert Birke\IEEEauthorrefmark{2}, Lydia Y. Chen%\thanks{$^2$ This work has been partly funded by the Swiss National
    \IEEEauthorrefmark{5} }
    \IEEEauthorblockA{\IEEEauthorrefmark{1}TU Delft, Netherlands
    z.zhao-8@tudelft.nl}    
    \IEEEauthorblockA{\IEEEauthorrefmark{2}ABB Research, Switzerland
    birke@ieee.org}
    \IEEEauthorblockA{\IEEEauthorrefmark{5}TU Delft, Netherlands
    lydiaychen@ieee.org}
}
\begin{document}
\xspaceremoveexception{-}
\title{\alg: Enhancing Table Synthesis via Fourier Transform}
\maketitle

\begin{abstract}
%\lc{I prefer networks instead of network}
Synthetic tabular data emerges as an alternative for sharing knowledge while adhering to restrictive data access regulations, e.g., European General Data Protection Regulation (GDPR). Mainstream state-of-the-art tabular data synthesizers draw methodologies from Generative Adversarial Networks (GANs), which are composed of a generator and a discriminator. While convolution neural networks are shown to be a better architecture than fully connected networks {for tabular data synthesizing}, two key properties of tabular data are overlooked: (i) the global correlation across columns, and (ii) invariant synthesizing to column permutations of input data.
To address the above problems, we propose a Fourier conditional tabular generative adversarial network (FCT-GAN). We introduce feature tokenization and Fourier {networks} to construct a transformer-style generator and discriminator, and capture both local and global dependencies across columns. The tokenizer captures local spatial features and transforms original data into tokens. Fourier networks transform tokens to frequency domains and element-wisely multiply a learnable filter. Extensive evaluation on benchmarks and real-world data shows that FCT-GAN can synthesize tabular data with high machine learning utility {(up to 27.8\% better than state-of-the-art baselines)} and high statistical similarity to the original data {(up to 26.5\% better)}, while maintaining the global correlation across columns, especially on high dimensional dataset. %Because 2D discrete Fourier transform and its inverse process do not involve any trainable parameter, therefore no parameter expansion when training data get bigger.
\end{abstract}

\section{Introduction}
While data sharing is crucial for knowledge development, privacy concerns and strict regulations (e.g., European General Data Protection Regulation (GDPR)) limit its full effectiveness. An emerging solution is to leverage synthetic data generated by machine learning models. Synthetic data has been powered by generative adversarial networks (GAN)~\cite{goodfellow2014generative} for various types of data, e.g., image~\cite{stylegan}, text to 
image~\cite{ramesh2021zero} and table~\cite{ctgan}.
% \rb{change table with something else? table is below}

Synthetic tabular data emerges as a prominent research direction because of its ample application scenarios in areas such as medicine~\cite{medgan} and finance~\cite{finance}. Compared to image data, one key difference of tabular data is that it is composed of different types of columns such as continuous, categorical or mixed variables. Therefore, GANs designed for image synthesis cannot be directly applied for tabular data. Previous works~\cite{ctgan,ctabgan,ctabplus} propose feature engineering solutions for different types of data such as using one-hot encoding for categorical variable. One-hot encoding is shown~\cite{ctgan} to better recover the categorical variable distribution for tabular GANs and capture inter-dependency across all the columns. However, one-hot encoding inevitably increases the data dimensions. High dimensional data\footnote{In this paper, dimension refers to the number of columns} is challenging for tabular GANs to learn global relations.
%\rb{unclear here why important, would move}. %The GAN algorithm designed for image synthesis, using convolutional neural network (CNN) as generator and discriminator~\cite{}, results into synthetic data better resembling the original data . \lc{Improve the connection}This is due to the fact that CNN can extract local spatial features well.
% \lc{Add CCN does not work for big data set. Be careful and clear about the definition of big data set. }
% \lc{be carecareful about the tradeoff off between the encoding and CNN limitation}
Prior studies~\cite{ctgan,ctabgan,ctabplus} show that the tabular GAN algorithms, which adopt CNNs as generator and discriminator, achieve better synthesis quality than using {purely} fully-connected neural networks. This is due to the fact that CNNs can extract local spatial features well. %\lc{I don't get why the following sentence} When using CNN for tabular GANs, rows are transformed into a fixed size image with columns being the pixels. %padding missing values with zeros.
The first limitation of directly adopting CNN to model tabular data is that it may overlook global relations between columns due to the size of the convolution filter. This limitation exacerbates when one-hot encoding is applied for categorical variables. Secondly, while permuting columns, e.g., reordering the columns by their types, does not have any semantic meaning, the local feature presentation extracted by convolution layers is distorted. 
%\lc{I don't see the connection so I can not rewrite.} 
%\lc{The explanation below is not easy to follow }
When using CNN for tabular GANs, one table row is transformed into one fixed-size image {by mapping each column value to a pixel}. The relationship between highly distant pixels, e.g. the pixel in the upper left corner and the pixel in the lower right corner, in a real image may not influence image classification. But for the tabular data wrapped as an image, these two pixels can represent highly correlated columns.
%\rb{should we add the problem of column explosion due to one-hot encoding?}

% \lc{Motivation example should be here}

To address the above two limitations, we propose
% a Fourier networks-based conditional tabular GAN. \lc{The wordings are slightly different from the abstract. In methodology, we call it block}
a conditional tabular GAN with Fourier Network \blocks (FNBs). The objective of the FNBs
is to learn the interactions among spatial locations in the frequency domain. We use FNBs for both the discriminator and generator with different designs. {The Fourier layer, which is the key part of an FNB, contains three operations: }
%The Fourier network architecture (shown in Fig.~\ref{fig:fourier_network}) is largely based on the vision transformers (ViT)~\cite{} \lc{why is this based on the vision transformer}. The Fourier network replaces the self-attention layer in ViT with the Fourier layer\rb{before we mention CNN, i.e. DNN with convolutional layers, while here we talk about attention layers. Better intro/flow needed.} which contains three key operations: 
(i) 2D discrete Fourier transform, (ii) element-wise multiplication between frequency-domain features and learnable weights and (iii) 2D inverse discrete Fourier transform. Furthermore, we process input data in a transformer-style tokenization way. A CNN-based filter is applied to original data to capture local spatial features and transform them into feature tokens. 
% For each row of tabular data, FN first wraps it as image and then break it into small parts (i.e., input tokens in ViT) as the input. 
Fourier layers transform tokens into frequency domain, then the learnable weights are applied to all the frequencies to learn the global relations. Our results show that \alg outperforms state-of-the-art (SOTA) up to 27.8\% in machine learning utility and 26.5\% in statistical similarity on 7 datasets. 
Thanks to Fourier \blocks ability to capture local and global relations, our results also show that among three different column orders, \alg has the least variation in synthesis quality among all comparisons. The experiment, with one high dimensional dataset, which 3 SOTA algorithms fail to train due to the data dimension issue, shows that \alg still maintains its performance and stability above all comparisons.

The main contributions of this study can be summarized as follows: %(1) This is the first study for analyzing the impact of column permutation on tabular data synthesizer. 
(1) We introduce the Fourier transform into tabular GAN training and design a generator and discriminator architecture. (2) Combining with a transformer-style input tokenizer, the novel architecture can capture both local and global relations of tabular data, leading to the desirable property of column permutation invariance. (3) We extensively evaluate FCT-GAN on 8 datasets against 5 state-of-the-art synthesizers, with a special focus on high dimensional real-world data.
% (1) Novel conditional adversarial network which introduces a classifier/regressor providing additional supervi- sion to improve the utility for ML applications. (2) Efficient modelling of continuous, categorical, and mixed variables via novel data encoding. (3) Improved GAN training us- ing well-designed information, downstream, generator losses along with Was+GP to enhance stability and effectiveness. (4) Constructed a simpler and more stable DP GAN algorithm for tabular data to control its performance under different privacy budgets.
% Fourier network learns the interactions among spatial locations in the frequency domains.

\section{Related Work}
We introduce various tabular data synthesizing methods and Fourier networks.
\subsection{Tabular Data Synthesizers}

% \lc{I don't see this paragraph fits into the related work. It more a problem statement or background. }
% To generate a realistic tabular data, there are two dimensions to consider: column-wise and row-wise. For each column, either the column be categorical, continuous or mixed, we expect the synthetic data approximates the data distribution of the real data. In row-wise, we expect the synthesizing algorithm can model the jointly probability distribution of all columns, so that the synthetic table maintains the same correlations among the columns.
% To generate a realistic tabular data, we expect: (i) 

The are various approaches for synthesizing tabular data.%, such as GANs, VAE\lc{Needs to introduced first}, etc. 
Probabilistic models such as Copulas~\cite{copulas} uses Copula function to model multivariate distributions. But categorical data can not be modeled by Gaussian Copula. Synthpop~\cite{nowok2016synthpop} works on a variable by variable basis by fitting a sequence of regression models and drawing synthetic values from the corresponding predictive distributions. Since it is variable by variable, the training process is computationally intense. Bayesian networks~\cite{privb,avino2018generating} are used to synthesize categorical variables. It lacks of the ability to generate continuous variables. % However, the data type limitation (e.g., only synthesize categorical variables) and computational problems prevent them to generate highly realistic data.

Current state-of-the-art introduces several tabular GAN algorithms. \tablegan~\cite{tablegan} introduces an auxiliary classification model along with discriminator training to enhance column dependency in the synthetic data. \ctgan~\cite{ctgan} and \ctab~\cite{ctabgan} improve data synthesis by introducing several preprocessing steps for categorical, continuous or mixed data types which encode data columns into suitable form for GAN training. The conditional vector designed by \ctgan and later improved by \ctab also helps the GAN training to reduce mode-collapse on minority categories. \ctabplus~\cite{ctabplus}, \pategan~\cite{pategan}, and \itgan~\cite{invgan} generate tabular data without risking privacy of original data by either adopting differential privacy or controlling the negative log-density of real records during the GAN training. However, as far as we know, there is no previous work studying tabular GAN algorithm which focuses on countering the influence of training data column permutation on final synthetic data quality.

% \lc{Add a puntchline of what the prior art did not do - permutation invariant}
\subsection{Fourier Networks}

The Fourier transform has played an important role in image processing for decades, e.g. JPEG compression~\cite{wallace92jpeg}. Incorporating Fourier transform into the neural network architecture design has been studied in many vision works~\cite{sitzmann2020implicit,chi2020fast,mathieu2013fast}. Recent work also leverages the  Fourier transform to design deep neural networks to solve partial differential equations (PDE)~\cite{li2020fourier} and NLP tasks~\cite{lee2021fnet}. Our Fourier network block architecture design is mainly inspired by the Global Filter Network (GFNet)~\cite{rao2021global}. We take the design of the input tokenization and global filter layer from GFNet and use it in our design of the generator and discriminator for tabular GAN.
% is largely based on the Global Filter Network (GFNet) qdesign but with some modifications. The GFNet is originally constructed for image classification and our objective is to design the generator and discriminator of tabular GAN.  
\section{Analysis on Permutation Invariance}
% \lc{Analysis sounds better than motivation. However, we need more results}
% \begin{figure}[t]
%     % \vspace{-0.7em}
% 	\begin{center}
% 		\subfloat[Avg-JSD variation]{
% 			\includegraphics[width=0.8\columnwidth]{LaTeX/figures/motivation_jsd.png}
% 			\label{fig:time_per_epoch}
% 		}
%  		\hfil
% 		\subfloat[Avg-WD variation]{
% 			\includegraphics[width=0.8\columnwidth]{LaTeX/figures/motivation_wd.png}
% 			\label{fig:varying_epochs}
% 		}
% % 		\vspace{-0.7em}
% 		\caption{Variation of statistical similarity difference between original and synthetic data with different column orders in training data.}
% 		\label{fig:timedistribution_intrusion}
% 	\end{center}
% % 	\vspace{-0.5em}
% \end{figure}

\begin{figure}
	\begin{center}
 			\includegraphics[width=0.8\columnwidth]{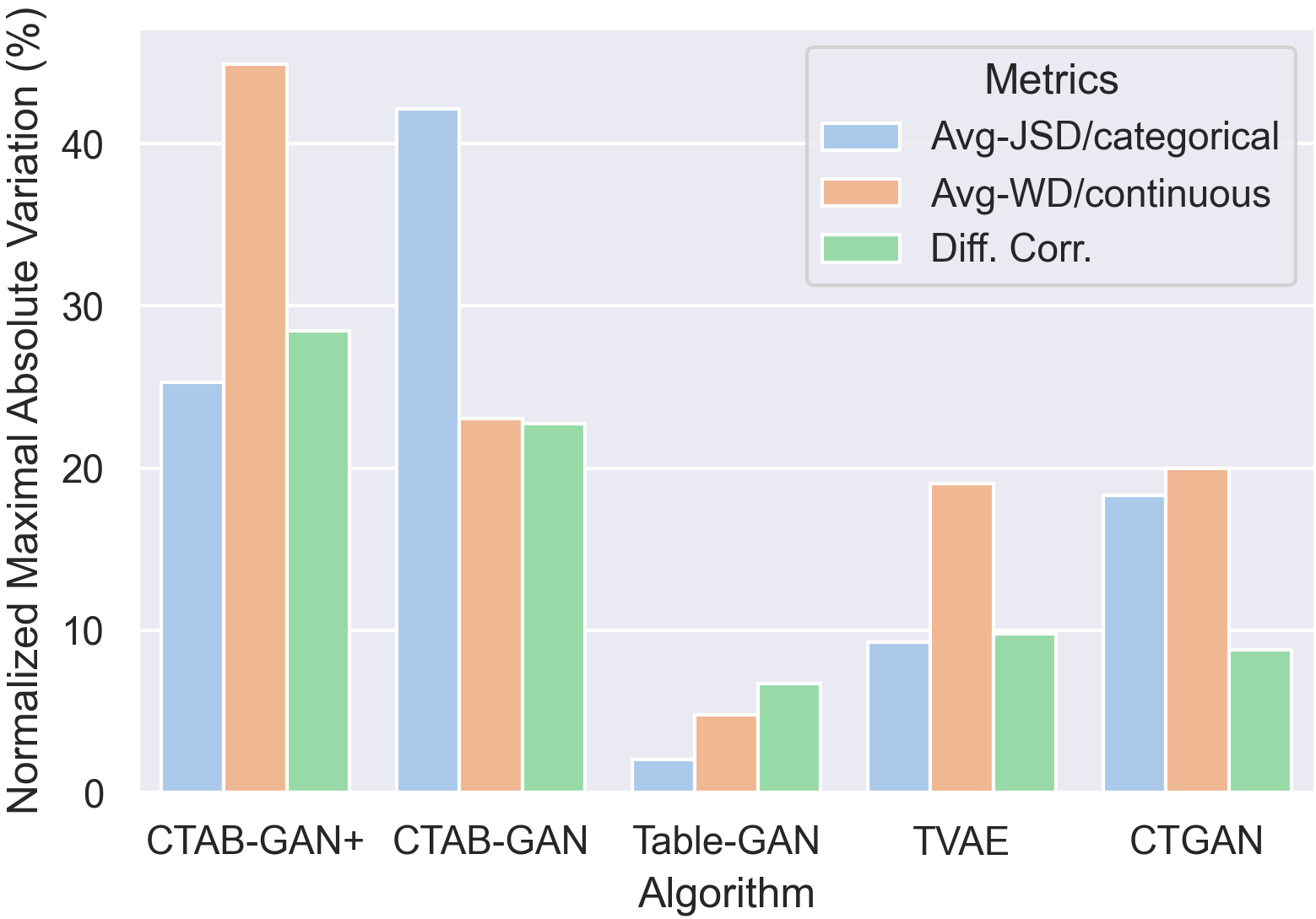}
% 			\missingfigure[width=0.99\textwidth]{Testing a long text string}
		\caption{Normalized maximal absolute variation (MAV) of difference in statistical similarity between original and synthetic data among different column orders: (1) Original, (2) Order by data type, and (3) Order by column correlation.}
		%\zz{wrong legend and tablegan} \lc{change to the modern presentation}} 
		\label{fig:mav_motivation}
 	\end{center}
 	\vspace{-1em}
\end{figure}
\begin{figure*}
	\begin{center}
			\includegraphics[width=0.75\textwidth]{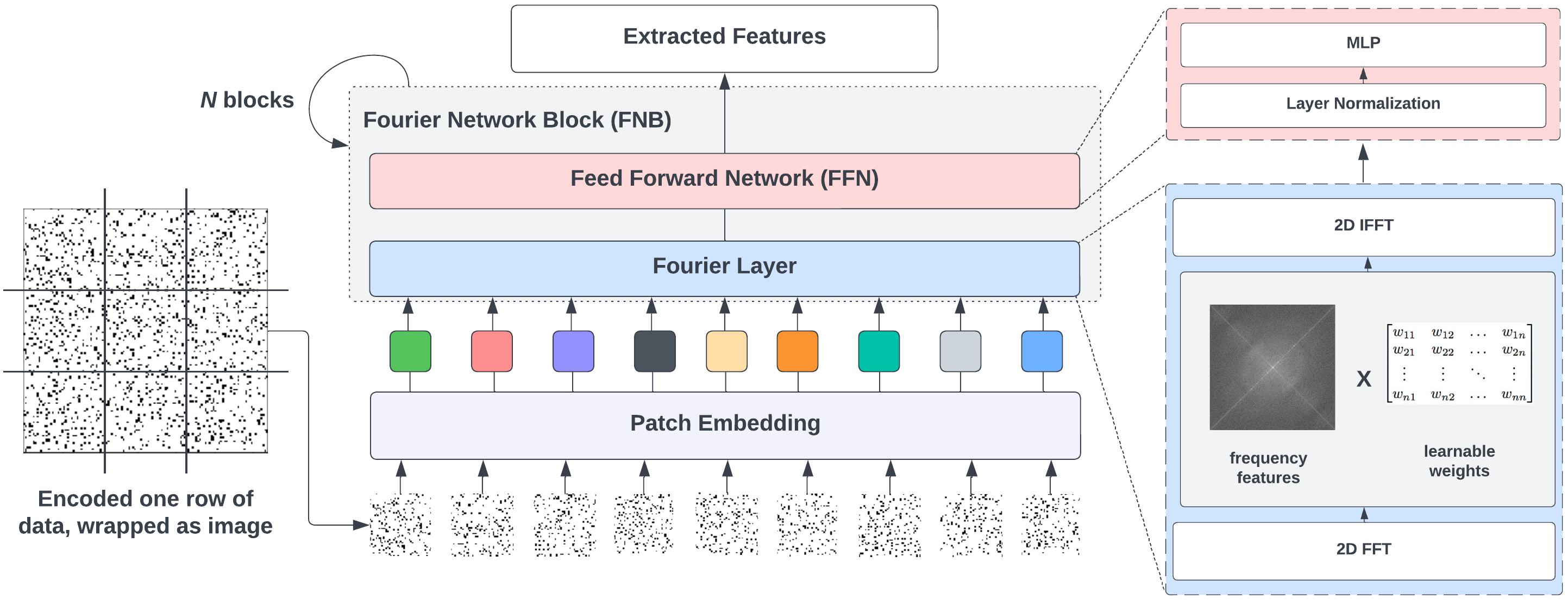}
% 			\missingfigure[width=0.99\textwidth]{Testing a long text string}
		\caption{Input Tokenization and Fourier Network Block. } 
		\label{fig:fourier_network}
 	\end{center}
 	\vspace{-1em}
\end{figure*}
\begin{table}[ht]
\centering
\caption{Average statistical similarity difference between original and synthetic data among three different column orders.}
\resizebox{0.85\columnwidth}{!}{
\begin{tabular}{ |c|c|c|c|c|c|c|c|c|c| }
\hline
\textbf{Method} & \textbf{Avg-JSD}& \textbf{Avg-WD} & \textbf{Diff. Corr.}   \\
\hline
{\ctabplus}  &\textbf{ 0.040}  & \textbf{0.012}  & \textbf{2.21}\\
\hline
  \ctab &\textbf{0.040} & 0.013&\textbf{2.21}\\\hline
 \tablegan    & 0.20 &0.017&3.57 \\\hline
  \tvae & 0.10 &0.231&2.79 \\\hline
 \ctgan     & 0067 & 0.038       &3.17 \\\hline
\end{tabular}
}
\label{table:permutation_analysis}
% \vspace{-0.8em}
\end{table}
We empirically demonstrate the instability of prior SOTA methods to permutations of the columns in the training data. Details on the datasets and evaluation metrics are provided in Sec.~\textbf{Experiment}.
% \rb{update with ref once available}.

We consider three column orders: (i) \textbf{Original}: as the name suggests, maintains the order as in the data downloaded from dataset source. (ii) \textbf{Order by data type}: puts all the continuous columns at the beginning and all categorical columns at the back. (iii) \textbf{Order by data correlation}: first calculates the pair-wise correlations between all columns. Then it sorts columns based on the absolute correlation value with highly correlated pairs in front and less correlated pairs later. Duplicate columns are skipped. We evaluate the statistical dissimilarity between real and synthetic table using Average Jensen–Shannon divergence (\textbf{Avg-JSD}) for all categorical columns and Average Wasserstein distance (\textbf{Avg-WD}) for all continuous columns. Finally, \textbf{Diff. Corr.} denotes the averaged distance between the correlation matrix of the real and the synthetic data. \textbf{The lower these metrics, the better the synthetic data quality}. 
For metric $\mathcal{M}$ and dataset $\mathcal{D}$, $\mathcal{L}$ = \{$V_{DM}^N$, $V_{DM}^N$ ... $V_{DM}^N$\} denotes the evaluation values on different column orders. We define Maximal Absolute Variation (MAV) as follows:
\begin{align*}
    \mbox{MAV} = max(\mathcal{L}) - min(\mathcal{L})
\end{align*}
MAV denotes the largest pair-wise difference among all order results. Note that different evaluation metrics can be on different scales. To have an easier comparison across all metrics for different algorithms, we further define the normalized MAV as follows:
\begin{align*}
    \mbox{normalized MAV (\%)} = \frac{\mbox{MAV}}{min(\mathcal{L})} \times 100
\end{align*}
% For metric $\mathcal{M}$ and dataset $\mathcal{D}$, $\mathcal{L}$ = \{$V_{DM}^N$, $V_{DM}^N$ ... $V_{DM}^N$\} denotes the evaluation values on different column orders, then we define Maximal relative variation (MRV)(\%) as follows:
% \begin{align*}
%     \mbox{MRV} = \frac{max(\mathcal{L}) - min(\mathcal{L})}{min(\mathcal{L})} \times 100
% \end{align*}

% Maximal relative variation (MRV) denotes the division between the largest pair-wise distance among all orders results and the smallest value among all the results.
%\rb{not clear, maye add formula either as ref or footnote?}. %For the 5 algorithms in Fig.~\ref{fig:mrv},
Fig.~\ref{fig:mav_motivation} presents the Normalized MAV of synthesis quality for 5 SOTA algorithms on 5 classification datasets (i.e., Loan, Adult, Intrusion, Credit and Covertype).
\ctab, \ctabplus and \tablegan use CNNs for generator and discriminator while \tvae and \ctgan use fully-connected networks. \tablegan is the only method that uses min-max normalization to encode all column types. The remaining four algorithms use one-hot encoding for categorical columns. This increases significantly the dimensions of the feature space making %es the encoded data much larger than the original one and thus 
it more difficult to learn global relations. Using CNN plus one-hot encoding makes \ctab and \ctabplus more vulnerable to column permutations. Fully-connected networks reduce this impact, but the variation is still non-negligible.  \tablegan is the most stable across all metrics, but its absolute performance is the worst among all SOTAs.  Tab.~\ref{table:permutation_analysis} presents the absolute results for each metric averaged over the same three column orderings. One can see that even though \ctab and \ctabplus experience some instability in their results, their performance ranges are still better than other algorithms.
% Therefore, they are still preferred tabular GAN algorithms\rb{not sure if this phrase is needed (next phrase is the main message}.
None of the SOTAs is able to provide tabular data synthesis both of high quality and invariant to column permutations.
 
% but later we will show that its absolute values are the worst among all SOTAs. None of the SOTAs is able to provide both high quality and column permutation invariant tabular data synthesis.
%invariant tabular synthesizer is needed.
\section{\alg}
In this section, we first present the primer of Fourier Network Blocks (FNB), and then describe the design of integrating them into the tabular GAN.

\begin{figure*}[t]
	\begin{center}
			\includegraphics[width=0.75\textwidth]{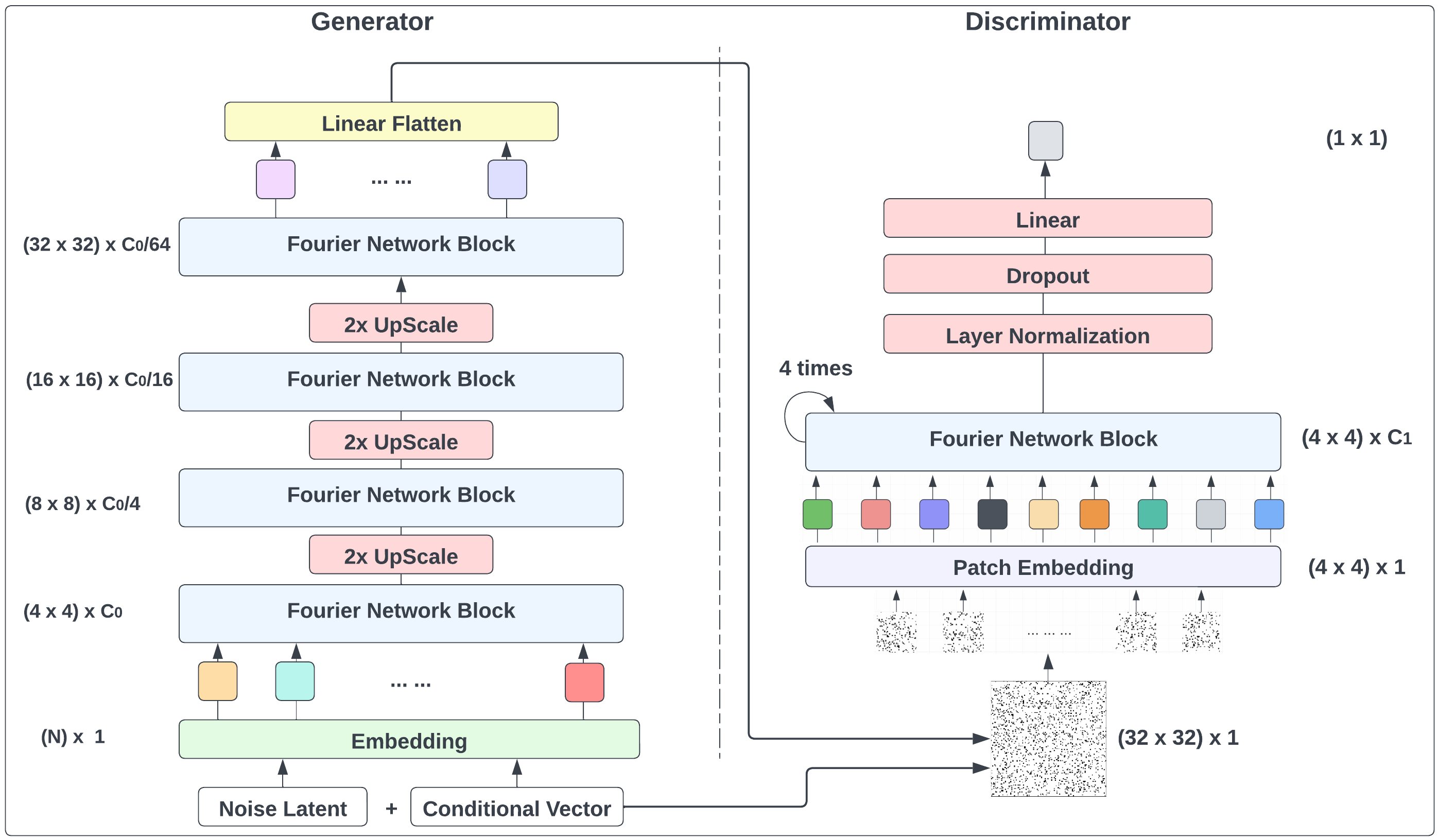}
% 			\missingfigure[width=0.99\textwidth]{Testing a long text string}
		\caption{Structure of \alg. Target generation dimension is setting to 32*32 as an example.  Due to page limitation, the auxiliary classifier/regressor model is omitted. }
		%\zz{(replace value by a symbol, emphasis the difference of two embedding in the figure, conditional vector captialize)}} 
		\label{fig:fctgan}
 	\end{center}
 	\vspace{-.7em}
\end{figure*}
\subsection{Primer on Fourier Network Blocks}
The input tokenization process and Fourier network block~\cite{rao2021global} are explored for image classification and can be combined in the following architecture, illustrated in Fig.~\ref{fig:fourier_network}. %This structure is largely based on GFNet~\cite{rao2021global} which is originally designed for image classification. 
This architecture is the base for constructing the generator and discriminator of \alg. 
%\rb{before going into details give a higher level description of the whole} \lc{Do so}.
{Each row of encoded tabular data is divided into tokens by a patch embedding layer. Then repeated layers of FNBs extract features from the tokens. In the generator the extracted features directly represent the generated data.
In the discriminator they are fed into a final layer for classification.
% The extracted features can be used as the generated data of generative model, or as the classification/regression features with further process.
} 

For input data tokenization (i.e., patch embedding in Fig.~\ref{fig:fourier_network}), we first break the original image into small parts to extract local spatial features (in the same vein as the ViT~\cite{vit}, MLP-Mixer~\cite{mlpmixer} and GFNet~\cite{rao2021global}). We define a CNN layer with a $k \times k$ kernel and stride the same as the kernel size. Therefore, for an image of size $N \times N$, after embedding, we have $\frac{N}{k} \times \frac{N}{k}$ tokens.

The Fourier Network Block consists of two parts: (1) Fourier layer and (2) Feed Forward Network. As suggested by GFNet, we use real fast Fourier transform (rFFT) (i.e., 2D FFT stage) in the Fourier layer since our input is a real tensor with no imaginary component.% \zz{(i.e., all the input data are real value without imaginary component)}.
Since the FFT of a real tensor is conjugate symmetric, we can cut off half of the frequency domain features without losing information. This helps to half the number of learnable weights in the frequency domain, hence accelerating computation.

The input tokenization and the FNB design can mix tokens representing different spatial locations allowing this architecture to take into account both local and global relations.

\subsection{Design of \alg}
Fig.~\ref{fig:fctgan} presents the overall structure of \alg. % We introduce the design of the generator and discriminator separately.
% \rb{Fig~\ref{fig:fourier_network} and Fig~\ref{fig:fctgan} are both very big and seem a bit redundant. Not sure how to change though}
% \lc{Add the high level description of key components and how they are connected to each other }
% \lc{Need to stress the conditional part}
We first introduce the data feature engineer method, then the network architectures of the generator and discriminator and how FNB is integrated.
% \lc{We need to explain the concept  of conditional GAN first}\rb{yes}
Finally, we discuss the training procedure and their respective loss functions.

\subsubsection{Data feature engineering} Before feeding tabular data into \alg, we adopt the data encoders from \ctabplus: one-hot encoding for categorical variables, and variational Gaussian mixtures (VGM) for continuous ones.
% For most of the continuous variables, variational gaussian mixture is estimated to encode the value.
For variables with specific distributions, e,g., uniform or normal, and for high dimensional categorical variables, min-max normalization is applied. Finally columns with missing values use a \textbf{Mixed-type encoder}~\cite{ctabgan}. 
\subsubsection{Conditional GAN}
% \lc{this should come up much earlier. The abstract emphasizes this condition tabular GAN a lot}\rb{agree} 
Due to the constraint on controlling generated data via GAN, conditional GAN is increasingly used. For each training batch, the same conditional vector is attached to noise latent as the input for generator, and also concatenate to real and generated data as the input for discriminator. Then the total input dimension for the generator and discriminator in conditional GANs is the sum of the original input plus conditional vector dimension.  The design of conditional vector of \alg is also adopted from \ctabplus, which indicates a specific category in the chosen categorical column or an exact Gaussian mode in the chosen continuous column if it uses variational Gaussian mixture to approximate the distribution.

\subsubsection{Generator.} The objective of the generator is to capture the joint probability distribution of all columns to synthesize high fidelity data. In \alg, we leverage the FNB to achieve the above goal. We opt for CNN-based GAN design which iteratively upscales the resolution at different stages. Hence our generator gradually increases the input sizes and reduces the embedding dimension at each stage.

Fig.~\ref{fig:fctgan} depicts an example generator design where the target resolution is $32 \times 32$ on the left. The noise latent and conditional vector are fed into an embedding layer (i.e., a Multi-Layer Perceptron which the input dimension is the same as the noise latent and conditional vector combined, the output dimension is $N$ = $H_0 \times W_0 \times C_0$) to convert them to a $H_0 \times W_0 \times C_0$ (by default we use $H_0$=$W_0$=4, $C_0$=256) vector. The vector is then reshaped into a $H_0 \times W_0$ resolution feature map with each point being a $C_0$-dimensional embedding. The result is fed into the first Fourier Network Block. 

After each FNB, we insert an upsampling module. There are several choices to achieve resolution-upscaling, such as bicubic interpolation or transpose convolutional operation. To mitigate memory usage and computation, we use the \textsc{pixelshuffle} function. This upscales the resolution of feature maps by a factor of 2$\times$ while reducing embedding dimension to a quarter. We repeat the FNB and UpScale stages until we reach the target resolution. The final linear unflatten layer is used to project the embedding into 1-dimension.

\subsubsection{Discriminator}
{Fig.~\ref{fig:fctgan} shows our discriminator architecture on the right. 
The objective of the discriminator is to distinguish real and fake data. We leverage repeated layers of Fourier network blocks from Fig.~\ref{fig:fourier_network}
to extract the features. Since \alg adopts the training structure of wasserstein GAN plus gradient penalty (WGAN+GP)~\cite{wgan_gp}, our discriminator outputs one single value instead of probability vector. This is achieved by a final group of normalization, dropout and linear layers.
%after extracting the features from FNB as shown in Fig.~\ref{fig:fourier_network}.
% feed the extracted features in Fig.~\ref{fig:fourier_network} to a normalization layer following a linear layer.
}
%This corresponds to a (binary) image classification task which can be solved by GFNet. Since \alg adopts the training structure of wasserstein GAN plus gradient penalty (WGAN+GP)~\cite{wgan_gp}, our discriminator will output one single value instead of probability vector. Therefore, we replace last two layers of GFNet (i.e., global average pooling and linear classifier) with our normalization and linear layers.}

{The generated data from the generator and the conditional vector are concatenated, wrapped as an image (padding missing values with zeros), and fed to the patch embedding layer of the discriminator. Within patch embedding, there is a CNN filter with a $k \times k$ kernel (we set $k=8$ by default), $C_1$ ($C_1=256$ by default) output channels, and stride same as the kernel size. Different from the generator, the inputs and outputs of the FNBs use constant dimensions. After 4 successive FNBs, the extracted features are flattened and downscaled to one single value.}

% \rb{this is not very clear. Especially relationship with GFNet and more detailed description needed)}
% The objective of the discriminator is to distinguish real and fake data. This corresponds to a (binary) image classification task which can be solved by GFNet. \rb{what is the exact relationship with GFNEt do we use the exact same structure?}
% % This goal is close to the GFNet which is original designed for image classification.
% Therefore, our discriminator is similar as the model shown in Fig.~\ref{fig:fourier_network}. The conditional vector and the generated data from the Generator are connected and wrapped as an image (padding missing values with zeros). Then a CNN filter is implemented to tokenize the image. 
% \zz{The kernel size of the CNN filter is set to 8 by default.}. Finally, the output features from last block are fed into a linear layer so that it outputs one single value in the end which better fits loss function using Wasserstein distance plus gradient penalty (WGAN+GP) \rb{ not sure if better here or where placed before}. \lc{objective function is usually described together with the architecture. This reason of having single value is lame}

\subsubsection{GAN training loss}
% \zz{for the loss, add the formula for G and D separately}
% Our discriminator outputs one single value instead of a probability vector (binary classification) is because we use  \cite{wgan_gp} and it demands that \rb{why does it demand that or rewrite sentence}.
% \lc{The discriminator loss and wasserstein distance and graident penalty need to be described as well}

% {\alg incorporates an auxiliary classifier/regressor as suggested in~\cite{tablegan, ctabgan, ctabplus} and adopts the basic GAN training loss from WGAN+GP ~\cite{wgan_gp}.
% This combines three training losses: (1) downstream loss, (2) information loss and (3) generator loss. }

% \alg adopts the basic GAN training loss from WGAN+GP ~\cite{wgan_gp}.
To improve the stability of GAN training, \alg adopts WGAN+GP~\cite{wgan_gp} loss. And to elevate the synthesizing performance, three extra training losses are added for generator: (1) downstream loss, (2) information loss and (3) generator loss.
% \rb{add overview of loss function, maybe with formula}
% \rb{The resulting loss function is ...} 
% which is added to generator loss and information loss.
\alg incorporates an auxiliary classifier/regressor as suggested in~\cite{tablegan, ctabgan, ctabplus}. For each synthesized data item the classifier/regressor outputs a predicted value using the synthesized features. The downstream loss quantifies the discrepancy between the synthesized and predicted value. This helps to increase the semantic integrity of synthetic records. The information loss matches the first-order (i.e., mean) and second-order (i.e., standard deviation) statistics of synthesized and real records. This leads to synthetic records with the same statistical characteristics as real records. The generator loss measures the difference between the given condition and the output class of the generator. This loss helps the generator learn to produce the exact same class as the given conditions. 

% The resulting losses for the generator (G) and the discriminator (D) are:}
% \begin{align*}
%   \mathcal{L}^{G} &= \mathcal{L}_{WGAN\_GP}^{G} + \mathcal{L}_{downstream}^{G} + \mathcal{L}_{info}^{G} + \mathcal{L}_{generator}^{G}  \\
%   \mathcal{L}^{D} &= \mathcal{L}_{WGAN\_GP}^{D}
% \end{align*}

% \lc{this should come up much earlier. The abstract emphasizes this condition tabular GAN a lot}\rb{agree} The design of conditional vector is also adopted from \ctabplus, which indicates a specific category in the chosen categorical column or an exact Gaussian mode in the chosen continuous column if it is with VGM.
\section{Experiment}
\label{sec:exp}
In this section, we evaluate the effectiveness of synthetic data generated by \alg in maintaining the global correlation and high statistical similarity to the original data which results in high quality downstream machine learning analyses. %experiments to answer two questions  (1) how much \alg improves synthetic data quality ? (2) to what extent \alg  mitigates the influence of column permutation in tabular training data? compared to the prior art.

\subsection{Experiment Setup}
{\bf Datasets}.  All algorithms are tested on 8 machine learning datasets. \textbf{Intrusion}, \textbf{Adult} and \textbf{Covertype} are from the UCI machine learning repository\footnote{\url{http://archive.ics.uci.edu/ml/datasets}}. \textbf{Credit} and \textbf{Loan} %\footnote{\url{https://www.kaggle.com/itsmesunil/bank-loan-modelling}}  
are from Kaggle\footnote{\url{https://www.kaggle.com/{mlg-ulb/creditcardfraud,itsmesunil/bank-loan-modelling}}}. These five tabular datasets are defined to have  a categorical variable as the target for conducting classification tasks. To include regression tasks we use two more datasets \textbf{Insurance}
 %\footnote{https://www.kaggle.com/mirichoi0218/insurance}
and \textbf{King} from Kaggle\footnote{\url{https://www.kaggle.com/{mirichoi0218/insurance,harlfoxem/housesalesprediction}}} where the target variable is continuous. Finally, \textbf{Youth}\footnote{The dataset is available from the first author upon request.} is a dataset from the Dutch government. It does not contain target variable. We include \textbf{Youth} because it contains many high dimensional categorical variables which lead to dimension explosion under one-hot encoding. This dataset is specially selected for testing algorithms' ability to capture global relations in high dimensional data and their stability to different training data column permutations.

% scalability\rb{or snesitivity to loval/global relations} of the algorithms.

 %The above five tabular datasets are used for classification tasks using as target a categorical variable. % for which we use the rest of the \columns to perform classification. 
%  To consider also regression tasks we use two more datasets, \textbf{Insurance}
%  %\footnote{https://www.kaggle.com/mirichoi0218/insurance}
%  and \textbf{King} from Kaggle\footnote{\url{https://www.kaggle.com/{mirichoi0218/insurance,harlfoxem/housesalesprediction}}} where the target variable is continuous.
 
Due to computing resource limitations, 50K rows of data are sampled randomly in a stratified manner with respect to the target variable for Covertype, Credit and Intrusion datasets.  %\lc{Be more precise on "maintaing the original distribution"} 
The Adult, Loan, Insurance, King and Youth datasets are taken in their entirety. The details of each dataset are shown in Tab.~\ref{table:DD}. We assume that the data type of each variable is known before training. 

For stability on column permutation analysis, three column orders are considered: (1) original, (2) order by data type and (3) order by data correlation which are already defined in section \textbf{Analysis on Permutation Invariance}.

% \cite{ctgan, ctabplus} holds the same assumption.

\begin{table}[ht]
\centering
\caption{Description of classification (C) and regression (R) Datasets. Con. and Cat. represent the number of continuous and categorical columns. Total Cat. represents the total number of categories across all categorical columns.}
\resizebox{1\columnwidth}{!}{
\begin{tabular}{ |c|c|c|c|c|c|c|c|c|c| }
\hline
\textbf{Dataset} & \textbf{Problem}& \textbf{Train/Test Split} & \textbf{$\mbox{Con.}$}  & \textbf{$\mbox{Cat.}$} & \textbf{$\mbox{Total Cat.}$}\\
\hline
{Adult}  & C.  & 39k/9k   &5 &9&104\\
\hline
  Covertype & C.& 40k/10k    &10 &45&94\\\hline
 Credit    & C.& 40k/10k    &1& 30&2\\\hline
  Intrusion & C.& 40k/10k    &22& 20&175\\\hline
 Loan     & C. & 4k/1k       &6&7&17\\\hline
 Insurance & R. &1k/300&3&4&14\\
 \hline
  King & R. &17k/4k&13&7&179\\\hline
  Youth & - &19k/5k&17&23&65774\\
\hline
\end{tabular}
}
\label{table:DD}
  \vspace{-0.5em}
\end{table}
\subsection{Experiment Metrics}

{\bf Baselines.} \alg is compared with 5 other SOTA tabular data synthesizing algorithms: \ctab, \ctabplus, \tablegan, \ctgan and \tvae. To separately show the utility of generator and discriminator, we also test two variants of \alg: \algg and \algd. \algg/\algd only keeps the generator/discriminator of \alg using as counterpart the CNN-based discriminator/generator %\algg/\algd adopts the design
from \ctabplus\footnote{These architectures are extracted from their offcial github https://github.com/Team-TUD/CTAB-GAN-plus.git.}. %\rb{does this give away too much for doubleblind?} \lc{I have the same concern} % which is the CNN-based model. 
% and \algd only keeps the discriminator of \alg.  
To have fair comparisons, all algorithms are implemented using Pytorch with hyper-parameters and network structures as set in the original papers. To ensure all algorithms converge, we train for 150 epochs on all datasets except Loan and Insurance trained for 300 and 500 epochs due to their small size. Each experiment is repeated 3 times and the average result is reported.

{\bf Environment}. Experiments run on a machine equipped with 32 GB memory, a GeForce RTX 2080 Ti GPU and a 10-core Intel i9 CPU under Ubuntu 20.04.

\subsection{Evaluation Metrics}
% \lc{We define MRV in section 2 but we never used it here}
% \lc{Grammar: only present tense. No future tense.}
The synthetic data is evaluated on two aspects: (1) machine learning utility and (2) statistical similarity.
\subsubsection{Machine learning utility}
Classification and regression datasets are quantified using different metrics, but they share the same evaluation process. 
%For all the algorithm,
We first train each algorithm on the training data and use the trained model to generate synthetic data of the same size as the training data. Then we use the training data and synthetic data to train same set of ML algorithms. For classification dataset, we choose decision tree classifier, linear support-vector-machine (SVM), random forest classifier, multinomial logistic regression and MLP. For regression dataset, we choose linear regression, ridge regression, lasso regression and Bayesian ridge regression model. Finally, we use the test set to separately test the two sets of ML models trained on the original and synthetic data. We use accuracy, F1-score and AUC as evaluation metrics for classification, and mean absolute percentage error (MAPE), explained variance score (EVS) and $R^2$ score as the metrics for regression. In the end, we calculate the difference between the results of the two sets of ML models for each metric. Since we report differences, the lower the result, the better the synthesis quality.

\subsubsection{Statistical similarity}
Three metrics are used to quantify the statistical similarity between real and synthetic data.

\textbf{Average Jensen-Shannon divergence (Avg-JSD)}. The JSD ~\cite{jsd} provides a measure to quantify the difference between the probability mass distributions of individual categorical variables belonging to real and synthetic data. This metric is bounded between 0 and 1 and is symmetric allowing for an easy interpretation of results.

\textbf{Average Wasserstein distance (Avg-WD)}. In a similar vein, the Wasserstein distance~\cite{wgan_test} is used to capture how well the distributions of individual continuous variables are emulated by synthetically produced data in correspondence to real data. 

We use WD because JSD metric is numerically unstable for evaluating the quality of continuous variables, especially when there is no overlap between the synthetic and original data. Hence, we resort to the more stable Wasserstein distance. 
 
\textbf{Difference in pair-wise correlation (Diff. Corr.)}. 
% To evaluate the preservation of column dependency in synthetic data, we compute the Pearson pair-wise correlation matrix for all columns within the real and synthetic datasets individually. We report the average absolute difference between the correlation matrices for real and synthetic data.
To evaluate the preservation of column dependency in synthetic data, we first compute the pair-wise correlation matrix for the columns within real and synthetic datasets individually. Pearson correlation coefficient is used between any two continuous variables. It ranges between $[-1,+1]$. Similarly, the uncertainty coefficient is used to measure the correlation between any two categorical features. It ranges between $[0,1]$. And the correlation ratio between categorical and continuous variables is used. It also ranges between $[0,1]$. Note that the dython\footnote{\url{http://shakedzy.xyz/dython/modules/nominal/\#compute_associations}} library is used to compute these metrics. Finally, the difference between pair-wise correlation matrices for real and synthetic datasets is computed.

\subsection{Result Analysis}
\begin{table*}[!ht]
\centering
\caption{Difference of ML Utility and Statistical Similarity between original and synthetic data. \textbf{ML Utility Difference R.} represents the results averaged on 2 regression datasets. \textbf{ML Utility Difference C.} represents the results averaged on 5 classification datasets. \textbf{ Statistical Similarity Difference} represents the results averaged on above 7 datasets. \textbf{Im. to 2nd Best} shows the improvement of \alg to the 2nd best result in the column excluding its own variants. Best results are on \textbf{bold}.}
\label{table:level}
\resizebox{0.97\textwidth}{!}{
\begin{tabular}{|c||c|c|c||c|c|c||c|c|c|}
\hline
\multirow{2}{*}{\textbf{Method}} & \multicolumn{3}{c||}{\bf ML Utility Difference R.} & \multicolumn{3}{c||}{\bf ML Utility Difference C.} & \multicolumn{3}{c|}{\bf Statistical Similarity Difference}\\
\cline{2-10}
 & \textbf{MAPE} & \textbf{EVS}  & \textbf{$R^2$} & \textbf{Accuracy} & \textbf{F1-score}  & \textbf{AUC} & \textbf{Avg-JSD} & \textbf{Avg-WD}  & \textbf{Diff. Corr.}\\
\hline
% \vspace{0.1em}
\small{\alg}   &\textbf{0.037}&\textbf{0.022}&\textbf{0.043} &\textbf{4.92\%}&\textbf{0.065}&\textbf{0.039}&\textbf{0.010}&	\textbf{0.034}&	\textbf{1.57}  \\
\small{\algd}   &0.042&0.041& 0.065&4.98\%&0.066&0.053&	0.014&	0.043&	1.72   \\
\small{\algg}  &0.131&0.041&0.109&  9.24\%&0.140&0.066&0.014&	0.047&	2.07 \\
\hline
\small{\ctabplus}  &\textbf{0.037}&0.025&\textbf{0.043}& {5.23}\% & {0.090} & {0.041} & 0.011 &	0.043&	1.65\\
\small{\ctab}    &0.059&0.594&0.707& 8.90\% & 0.107 &   0.094  & 0.021 &	0.056&	1.70 \\
\small{\ctgan}    &0.871&0.594&0.709& 21.51\% & 0.274 &   0.253 & 0.037&	0.090&	2.96   \\
\small{TVAE}    &0.243 &0.078&0.215&11.11\%&0.100 &0.230&0.025&	0.122&	2.41   \\
\small{\tablegan} &0.338&0.434&0.479& 11.40\%  & 0.130 &   0.169 & 0.028&	0.231&	3.23  \\
\hline
\hline
Im. to 2nd Best &0\%&13.6\%&0\%&6.3\%&27.8\%&5.1\%&10\%&26.5\%&5.1\%\\
\hline
\end{tabular}
}
%\vspace{-1em}
\end{table*}

\begin{table*}
\centering
\caption{Maximal absolute variation (MAV) of the difference of ML Utility and Statistical Similarity on three type of column orders between original and synthetic data. \textbf{ML Utility Difference R.} represents the results averaged on 2 regression datasets. \textbf{ML Utility Difference C.} represents the results averaged on 5 classification datasets. \textbf{ Statistical Similarity Difference} represents the results averaged on above 7 datasets. Best result are on \textbf{bold}}
\label{table:variation}
\resizebox{0.97\textwidth}{!}{
\begin{tabular}{|c||c|c|c||c|c|c||c|c|c|}
\hline
\multirow{2}{*}{\textbf{Method}} & \multicolumn{3}{c||}{\bf ML Utility Difference R.} & \multicolumn{3}{c||}{\bf ML Utility Difference C.} & \multicolumn{3}{c|}{\bf Statistical Similarity Difference}\\
\cline{2-10}
 & \textbf{MAPE} & \textbf{EVS}  & \textbf{$R^2$} & \textbf{Accuracy} & \textbf{F1-score}  & \textbf{AUC} & \textbf{Avg-JSD} & \textbf{Avg-WD}  & \textbf{Diff. Corr.}\\
\hline
\small{\alg}    &\textbf{0.04}&\textbf{0.02}&\textbf{0.03}&\textbf{0.02}&\textbf{0.01}&\textbf{0.01}&3e-3&2e-3&\textbf{0.11}  \\
\small{\algd}    &0.11 &0.03&0.2 &\textbf{0.02}&0.03&\textbf{0.02}& 	7e-3&5e-3&	0.2 \\
\small{\algg}     & 0.08&0.11&0.08&0.03&0.05&0.03  &9e-3&\textbf{1e-3}&	0.2\\
\small{\ctabplus}  &0.28   &0.26    &0.35 &\textbf{0.02}&0.03&0.03&5e-3&4e-3&0.32\\
\small{\ctab}    &0.05&0.56&0.67&0.07 &0.17&0.12 &6e-3&7e-3&0.3\\
\small{\ctgan}    &0.57&0.25&0.38&0.09&0.06&  0.09&0.01&0.01&0.17\\
\small{TVAE}   &0.15&0.03&0.05&0.04&0.05&0.05 &\textbf{2e-3}&3e-3&0.15 \\
\small{\tablegan}  &0.09&0.3&0.16&0.14&0.24&0.1 	&0.01 &2e-3&0.28\\
\hline
\end{tabular}
}
% \vspace{-0.3em}
\end{table*}

\subsubsection{Quality of synthetic data}
We first analyze the algorithm performance on the classification and regression datasets. Then, we present results on \textbf{Youth}, a larger real world dataset.

Tab.~\ref{table:level} shows the ML utility and statistical similarity results of 8 algorithms on 7 datasets. It is worth noting that as results summarize the difference between real and synthetic data, the lower the value the better the result. The table is grouped into three sections: averaged ML Utility results across regression datasets, averaged ML utility for classification datasets and averaged statistical similarity on all 7 datasets. Best results are highlighted in bold.
% Columns 2 to 4 are only the ML utility result averaged on the regression dataset. Columns 5 to 7 are only dedicated for classification dataset result. Columns 7 to 10 report the statistical similarity result averaged on all above 7 datasets.
One can see that \alg outperforms all the baselines on all the metrics. Excluding its direct own variants, \alg also significantly outperforms the second best algorithm, i.e., CTAB-GAN+, in most metrics. Note that many of our classification datasets have an uneven class distribution, then F1-score is a more appropriate metric than accuracy.
%to be focused.
The results shows that \alg improves \ctabplus by 27.8\% on F1-score, which shows that the new architecture enhances the ability to capture global column dependency. Among \alg and its variants, \algd shows the performance relatively close to \alg. \algg is still better than most of SOTAs except \ctabplus, but it degrades more from \alg compared to \algd. This shows that the generator can not fully demonstrate its ability without a well-designed discriminator, but a stronger discriminator can level up the synthesizing ability of the generator by better identifying fake data and hence forcing the generator to improve the fidelity of the generated data.  

\subsubsection{Stability on column permutation}
We show the maximal absolute variation of each metric on all algorithms among three column orders in Tab.~\ref{table:variation}. \alg achieves the smallest MVA in 7 out of 9 metrics and is the second best on the other 2 metrics. This means that no matter which column order is used, \alg can always capture the (global and local) column dependencies better than any other baseline. As \tvae uses fully-connected network layers only for both generator and discriminator, it also shows better stability. The same conclusion does not hold for \ctab even if it also uses only fully-connected networks. We can conclude that fully-connected networks can help some of the generative models to counter the influence of column permutations on the quality of the synthesized data, but it is not guaranteed. 
%\lc{The following should be highlighted earlier. Table shows the ML uitlity first}
Moreover one can note that changing training data column order generally influences more ML utility than statistical similarity. This means that for most of the tabular GAN algorithms, it is more difficult to capture the, especially global, column dependencies rather than to model the distribution of every single variable.

\subsubsection{Results on Youth dataset}
The analysis on the Youth dataset is singled out because it contains many categorical columns with large number of  categories. Encoding these columns with one-hot encoding  translates into a high number of dimensions, i.e., 65774 (see Tab.~\ref{table:DD}).
% , we know it contains 65774 categories in total. If we only use one-hot encoding for representing each category, the data dimension explodes. 
Moreover, the algorithms such as \ctgan and \ctab are  conditional GANs which require an additional conditional vector. This exacerbates even more the dimensionality issue. To construct the conditional vector for these categories, we need another 65774 dimension vector so that we can indicate each of the category.
%the input data (i.e., encoded data + conditional vector) for discriminator is more than 13k \rb{not clear where 13K comes from?}  columns.
%for only considering the categorical data encoding.
This stresses the resource demands and we were not able to train \textbf{\tvae, \ctgan and \ctab on Youth dataset even with batch size equal to 1.} in our compute environment. 
%\lc{does the following imply that we use the same feature engineering as ctabganplus}\rb{to my undertandning yes, see previous section} 
Using our feature engineering, we can encode some high dimensional categorical columns by min-max normalization lowering the resource footprint. % and so does \ctabplus. 

\begin{table}[ht]
\centering
\caption{Statistical similarity difference on \textbf{Youth} dataset. MAV in the parenthese. \ctab \ctgan and \tvae can not train on this dataset due to the huge dimension of encoded training data.}
\resizebox{1\columnwidth}{!}{
\begin{tabular}{ |c|c|c|c|c|c|c|c|c|c| }
\hline
\textbf{Method} & \textbf{Avg-JSD}&  \textbf{Avg-WD} &  \textbf{Diff. Corr.} & \textbf{ED Dim.} & \textbf{CV Dim.} \\
\hline
{\alg}  &	\textbf{0.42} (\textbf{0.011})&\textbf{0.002} (\textbf{5e-4})& \textbf{8.74} (\textbf{0.21}) & 376& 343\\\hline
  \algd &	0.42 (0.018) & 0.002 (2.5e-3)&	9.11 (0.33)&376& 343\\\hline
 \algg    &	0.42 (0.04)& 0.004 (6e-4)&	9.46 (0.34)&376& 343\\\hline
  \ctabplus 	&0.42 (0.037)& 0.01 (2.6e-3)&	9.80 (0.33)&376& 343\\\hline
 \tablegan    &	0.48 (0.02) &0.006 (8e-4)  &	9.64 (0.27)&40& -\\\hline
\end{tabular}
}
\label{table:youth}
\vspace{-.6em}
\end{table}

Tab.~\ref{table:youth} summarizes the result on all algorithms that can still train on the Youth dataset. The \textbf{ED Dim.} is the encoded data dimension of training data. This is also the dimension of the data generated at generator output. The \textbf{CV Dim.} represents the conditional vector dimension.
% \lc{this part should be mentioned already in the method}
% The total input dimension for the discriminator in conditional GANs is the sum of the two: encoded data plus conditional vector dimension. 
The first 4 algorithms in Tab.~\ref{table:youth}, generate data that is $32 \times 32$ (i.e., upscale from 376 to $k\times k$ 
where k is an integer power of 2) and the input dimension of the discriminator is  $32 \times 32$ (i.e., upscale from 376+343=719 to $k\times k$). Results show that \alg outperforms all baselines on both synthesis quality and stability. The difference between \alg and \ctabplus on \textbf{Diff. Corr.} demonstrates that CNN-based GANs can not capture global relations well with increasing dimensions of data. Though \tablegan is a CNN-based solution, its min-max normalization does not change the data dimension and outperforms \ctabplus on \textbf{Diff. Corr.}. That is because it operates on a smaller scale image ($8 \times 8$) which makes it easier to capture the pixel (column) dependencies. Still, its overall performance trails far behind \alg.

\section{Conclusion}
In this paper, we present a novel tabular GAN algorithm -- \alg. \alg leverages the transformer-style tokenizer and Fourier network blocks to design the generator and discriminator. For discriminator, the tokenizer uses a CNN-based filter to extract local spatial features from original input data and tokenizes them for Fourier network blocks. The key component of Fourier network blocks -- the Fourier layer uses 2D FFT/IFFT to transform input tokens into frequency domain and applies learnable filters on the frequency features to ease learning global relations. This design takes both local and global features into account. The generator imitates the design of CNN-based GANs, which piles up Fourier network blocks and upscales feature dimensions at each layer by 2$\times$ until reaching the target size. The proposed method surpasses state-of-the-art performance, especially on high dimensional data. It also shows brilliant stability to counter the impact of training data column permutations on the synthetic data quality.
%\newpage
\bibliographystyle{abbrv}
\bibliography{main}
\end{document}